\documentclass[conference]{IEEEtran}
\usepackage{blindtext, graphicx}

\usepackage[utf8]{inputenc}
\usepackage[english]{babel}
\usepackage{cite}
\usepackage{url}

\usepackage{float}

\usepackage{hyperref}
\hypersetup{
    colorlinks=true,
    linkcolor=red,
    filecolor=magenta,      
    urlcolor=cyan,
    citecolor=red
}
 
\usepackage[framemethod=TikZ]{mdframed}

\mdfdefinestyle{MyFrame}{%
    frametitle={\colorbox{white}{\space##1\space}},
    linecolor=black,
    outerlinewidth=1pt,
    roundcorner=10pt,
    innertopmargin=8pt,
    frametitleaboveskip=-1.4\ht\strutbox, %
    frametitlealignment=\center,
    innerbottommargin=\baselineskip,
    innerrightmargin=20pt,
    innerleftmargin=10pt,
    backgroundcolor=blue!10}

\mdfdefinestyle{irati}{%
    frametitle={\fcolorbox{white}{\space##1\space}},
    linecolor=white,
    outerlinewidth=0pt,
    roundcorner=0pt,
    innertopmargin=8pt,
    frametitleaboveskip=-1.4\ht\strutbox, %
    frametitlealignment=\center,
    innerbottommargin=\baselineskip,
    innerrightmargin=20pt,
    innerleftmargin=10pt,
    backgroundcolor=gray!10}

\ifCLASSINFOpdf
\else
\fi
\usepackage{algorithmic}
\usepackage{algorithm}
\usepackage[normalem]{ulem}

\usepackage[most]{tcolorbox}

\tcbset{
    frame code={}
    center title,
    left=0pt,
    right=0pt,
    top=0pt,
    bottom=0pt,
    colback=blue!10,
    colframe=white,
    width=0.5\textwidth,
    enlarge left by=0mm,
    boxsep=5pt,
    arc=0pt,outer arc=0pt,
    }

\begin{document}
%






\title{Dissecting Robotics --- historical overview and future perspectives}

\author{Irati Zamalloa, Risto Kojcev, Alejandro Hernández, \\ Iñigo Muguruza, Lander Usategui, Asier Bilbao and Víctor Mayoral \\ 
\\
\emph{Acutronic Robotics, April 2017} 
\\}


%


\maketitle
\thispagestyle{plain}
\pagestyle{plain}


\begin{abstract}


Robotics is called to be the next technological revolution and estimations indicate that it will trigger the fourth industrial revolution.
This article presents a review of some of the most relevant milestones that occurred in robotics over the last few decades and future perspectives. Despite the fact that, nowadays, robotics is an emerging field, the challenges in many technological aspects and more importantly bringing innovative solutions to the market still remain open. The need of reducing the integration time, costs and a common hardware infrastructure are discussed and further analysed in this work. We conclude with a discussion of the future perspectives of robotics as an engineering discipline and with suggestions for future research directions.
\end{abstract}


\begin{IEEEkeywords}
robotics, review, hardware, H-ROS, artificial intelligence.
\end{IEEEkeywords}

%
\IEEEpeerreviewmaketitle


\section{Introduction}

The first use of the word \emph{"Robot"} dates back in 1921 and it was introduced by Karel Čapek in his play \emph{Rossum's Universal Robots}. The play describes mechanical men that are built to work on the factory assembly lines and that rebel against their human masters \cite{low2007industrial}. The etymological origin of the word Robot is from the Czech word \emph{robota}, which means servitude or forced labor.\\
\newline
The term \emph{"Robotics"} was first mentioned by the Russian-born American science-fiction writer Isaac Asimov in 1942 in his short story \emph{Runabout} \cite{low2007industrial}. Asimov had a much brighter and more optimistic opinion of the robot's role in human society compared to the view of Capek. In his short stories, he characterized the robots as helpful servants of man.  Asimov defined robotics as the science that study robots. He created the \textbf{Three Laws of Robotics}, that state the following:

\begin{itemize}
    \item \emph{The First Law} states that a robot should not harm a person or let a person suffer damage because of their inaction.
    \item \emph{Second Law} states that a robot must comply with all orders that a human dictates, with the proviso that occurs if these orders were in contradiction with the First Law.
    \item \emph{The Third Law} states that a robot must take care of its own integrity, except when this protection creates a conflict with the First or Second Law.
\end{itemize}

\noindent Over the last decades, robotics evolved from fiction to reality becoming the science and technique that is involved in the design, manufacture and use of robots. Computer Science, Electrical Engineering, Mechanical Engineering and Artificial Intelligence (AI) are just some of the disciplines that are combined in robotics. The main objective of robotics is the construction of devices that perform user-defined tasks. The rapid growth of the field in scientific terms had led the development of different types of robots. Examples of different robotic systems are: industrial robots, manipulators, terrestrial, aerial, aquatic, research, didactic, entertainment robots or humanoids.

\noindent Section \ref{Robotics_Evolution} covers the evolution of robotics over the last decades including current trends. In Section \ref{h_ros}, we present the future perspectives of robotics based on the historical overview presented in \ref{Robotics_Evolution}. In Section \ref{discussion} we conclude this work and discuss our viewpoint for future development in robotics.

\begin{figure*}
\centering
\includegraphics[width=1\textwidth]{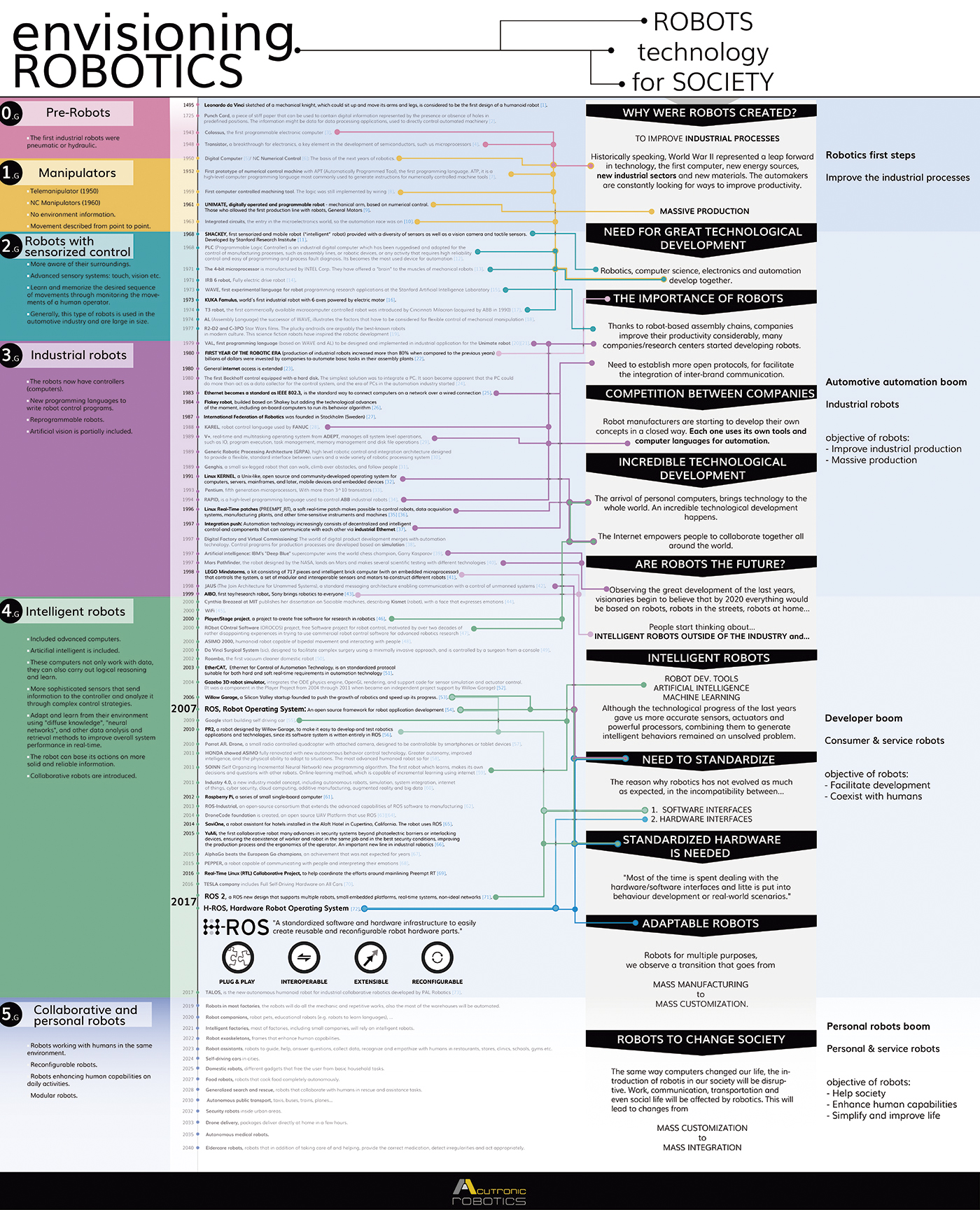}
\caption{A historical overview of the most relevant milestones that occurred in robotics over the last decades. The timeline also provides an insight and the vision of Erle Robotics for the future of robotics. \textcopyright Acutronic Robotics 2017. All Rights Reserved. A high resolution image is available at \cite{erleEnvisioning}}
\label{timeline}
\end{figure*}


\section{The evolution of Robotics}
\label{Robotics_Evolution}

Figure \ref{timeline} illustrates a historical overview of the growth of robotics, divided into four columns: the first column presents the different robot generations, the second column (central) shows some of the most relevant milestones and our vision of the upcoming future of robotics,  the third one the needs and social impact of robotics, and the fourth summarizes the different stages of development in robotics.\\
\newline
The progress of robotics is influenced by the technological advances, for example, the creation of the transistor \cite{transistor}, the digital computer \cite{506395}, the numerical control system \cite{middleditch1973survey} or the integrated circuits \cite{saxena2009invention}. These technological advances further enhanced the properties of the robots, and helped them evolve from solely mechanical or hydraulic machines, to programmable systems, which can even be aware of their environment. Similar to other technological innovations, robotics has advanced and changed taking into account the needs of society.


\noindent Based on their characteristics and the properties of the robots, we can classify the development of robotics into four generations: 

\subsection{\textbf{Generation 0: Pre-Robots (up to 1950)}}

\begin{tcolorbox}[colback=gray!10]
\textbf{Characteristics:}\\
\begin{itemize}
    \item The first industrial robots were pneumatic or\\
    hydraulic.\\
\end{itemize}
\end{tcolorbox}

In 1495, the polymath Leonardo Da Vinci envisioned the desing of the first humanoid \cite{firsthumanoid}. In the following years, different machines were manufactured using mechanical elements that helped the society and the industry. It was not until the first industrial revolution that factories began to think about automation as a way of improving manufacturing processes. The automated industrial machines of this generation were based on pneumatic or hydraulic mechanisms, lacking of any computing capacity and were managed by the workers.
\newline 
The first automation techniques were the punch cards \cite{punchcard}, used to enter information to different machinery (e.g. for controlling textile looms). The first electronic computers, for example the Colossus \cite{colossus}, also used punched cards for programming. 


\subsection{\textbf{Generation 1: First Manipulators (1950-1967)}}
\label{generation1}

\begin{tcolorbox}[colback=gray!10]
\textbf{Characteristics:}\\
\begin{itemize}
    \item Lack of information regarding the environment.
    \item Simple control algorithms (point-to- point).
\end{itemize}
\end{tcolorbox}

Due to the rapid technological development and the efforts to improve industrial production, automated machines were designed to increase the productivity. The machining-tool manufacturers introduced the numerical controlled (NC) machines which enabled other manufacturers to produce better products \cite{middleditch1973survey}. The union between the NC capable machining tools and the manipulators paved the way to the first generation of robots.\\
\newline
Robotics originated as a solution to improve output and satisfy the high quotas of the U.S. automotive industry. In parallel the technological growth led to the construction of the first digitally controlled mechanical arms which boosted the performance of repeatable, "\emph{simple}" tasks such as pick and place. The first acknowledged robot is UNIMATE \cite{unimate} (considered by many the first industrial robot), a programmable machine created by George Devol and Joe Engleberger that two years before funded the world's first robot company called Unimation (Universal Automation). In 1960, they secured a contract with General Motors to install the robotic arm in their factory located in Trenton (New Jersey). UNIMATE helped improve the production, which further motivated many companies and research centers to actively dedicate resources in robotics. 

\subsection{\textbf{Generation 2: Sensorized robots (1968-1977)}}

\begin{tcolorbox}[colback=gray!10]
\textbf{Characteristics:}\\
\begin{itemize}
    \item More awareness of their surroundings.
    \item Advanced sensory systems: for example, force, torque, vision.
    \item Learning by demonstration. 
    \item These type of robots are used in the\\automotive industry and have large footprint.\\
\end{itemize}
\end{tcolorbox}

Starting from 1968, the integration of sensors marks the \emph{second generation of robots}. These robots were able to react to the environment and offer responses that met different challenges. Shakey \cite{shakey}, developed by Stanford Research Institute, was the first sensorized mobile robot, containing a diversity of sensors (for example tactile sensors) as well as a vision camera.\\
\newline
During this period relevant investments were made in robotics. In the industrial environment, we have to highlight the PLC (Programmable Logic Controller) \cite{bryan1988programmable}, an industrial digital computer, which was designed and adapted for the control of manufacturing processes, such as assembly lines, robotic devices or any activity that requires high reliability. PLCs were at the considered to be easy to program. Due to these characteristics, PLCs became a commonly used device in the automation industry.\\
\newline
In 1973, KUKA (one of the world's leading manufacturers of industrial robots) built the first industrial robot with 6 electromechanical-driven axes called Famulus \cite{kuka}. One year later, the T3 robot \cite{cutkosky1982position} was introduced in the market by Cincinnati Milacron (acquired by ABB in 1990). The T3 robot was the first commercially available robot controlled by a microcomputer.


\subsection{\textbf{Generation 3: Industrial robots (1978-1999)}}
\label{industrialrobots}

\begin{tcolorbox}[colback=gray!10]
\textbf{Characteristics:}\\
\begin{itemize}
    \item Robots now have dedicated controllers (computers).
    \item New programming languages for robot control.
    \item Reprogrammable robots.
    \item Partial inclusion of artificial vision.\\
\end{itemize}
\end{tcolorbox}

Many consider that the Era of Robots started in 1980 \cite{roboticsera}. Billions of dollars were invested by companies all around the world to automate basic tasks in their assembly lines. The investments in automation solutions increased the sales of industrial robots up to 80\% compared to previous years. Robots populated many industrial sectors to automate a wide variety of activities such as painting, soldering, moving or assembly.\\
\newline
Key technologies that still drive the development of robots appeared during these years: general Internet access was extended in 1980 \cite{internet}, Ethernet became a standard in 1983 \cite{ethernet} (IEEE 802.3), the Linux kernel was announced in 1991 \cite{linux} and soon after, real-time patches started appearing \cite{linuxreal,Yodaiken_cheapoperating} to increase the determinism of Linux-based systems.\\
\newline
The \emph{“robot programming languages”} also became popular during this time. For example, Unimation started using \emph{VAL} in 1979 \cite{val} \cite{val1}, FANUC designed \emph{Karel} in 1988 \cite{fanuc} and in 1994 ABB created \emph{Rapid} \cite{abb} making robots re-programmable machines which also contained a dedicated controller.\\
\newline
By the end of the 1990s, companies started thinking about robots outside industrial environments. Among the robots and kits created within this period, we highlight two that became an inspiration for hundreds of roboticists:

\begin{itemize}

    \item \textbf{The first LEGO Mindstorms kit (1998)} \cite{LEGO}, a set consisting of 717 pieces, including LEGO bricks, motors, gears, different sensors, and a RCX Brick with an embedded microprocessor to construct various robots using the same parts. The kit allowed to teach the principles of robotics. Creative projects have appeared over the years showing the potential of interchangeable hardware in robotics. 

    \item \textbf{Sony’s AIBO (1999)} \cite{AIBO}, the world’s first entertainment robot. Widely used for research and development. Sony brought robotics to everyone with a \$1,500 robot that included a distributed hardware and software architecture. The OPEN-R architecture involved the use of modular hardware components —e.g. appendages that can be easily removed and replaced to change the shape and function of the robots—, and modular software components that can be interchanged to modify their behavior and movement patterns. OPEN-R represented an inspiration for future robotic frameworks and showed promise to minimize the need for programming individual movements or responses.
\end{itemize}

\noindent Sony’s AIBO and LEGO’s Mindstorms were built upon the principle of modularity, both concepts were able to easily exchange components and both of them presented common infrastructures. Even though they came from the consumer side of robotics, one could argue that their success was strongly related to the fact that both products made use of interchangeable hardware and software modules. The use of a common infrastructure proved to be one of the key advantages of these technologies. 

\subsection{\textbf{Generation 4: Intelligent robots (2000-2017)}}
\label{Intelligent_Robots}

\begin{tcolorbox}[colback=gray!10]
\textbf{Characteristics:}\\
\begin{itemize}
    \item Inclusion of advanced computing capabilities.
    \item These computers not only work with data,\\they can also carry out logical reasoning\\and learn.
    \item Artificial Intelligence begins to be\\ included partially and experimentally.
    \item More sophisticated sensors that send informa-\\tion to the controller and analyze it through\\complex control strategies.
    \item The robot can base its actions on more solid\\and reliable information.
    \item Collaborative robots are introduced.\\
\end{itemize}
\end{tcolorbox}

The fourth generation of robots, dating from the 2000, consisted of more intelligent robots that included advanced computers to reason and learn. These robots also contained more sophisticated sensors that helped them adapt more effectively to different circumstances.\\
\newline
The robot Roomba \cite{jones2005autonomous} —the first vacuum cleaner domestic robot— introduced the robots in many homes. YuMi \cite{yumi}, the first collaborative robot included many advances in the security systems beyond photoelectric barriers or interlocking devices, ensuring the coexistence of worker and robot in the same environment, improving the production process and the ergonomics of the operator. These advances both on the human-robot collaboration and improvements of the robot security systems, have allowed the robots to work together with humans in the same environment.\\
\newline
Among the technologies that appeared in this period we highlight the Player Project \cite{player} (2000, formerly the Player/Stage Project), the Gazebo simulator \cite{gazebo} (2004) and the Robot Operating System \cite{ROS} (2007). Moreover, relevant hardware platforms appeared during these years. Single Board Computers (SBCs) like the Raspberry Pi \cite{RPi} enabled millions of users all around the world to easily create robots.\\

The following subsections will describe some relevant  observations our research obtained within this period:

\subsubsection{\textbf{Decline in Industrial robot innovation}}
\label{IndustrialRobots}
Except for the appearance of the collaborative robots in 2015, the progress within the field of industrial robotics has significantly slowed down compared to previous decades. While industrial robots significantly improved their accuracy, speed or offer greater load capacity, industrial robots regarding innovation up till today are very stagnant.

\subsubsection{\textbf{The boost of bio-inspired Artificial Intelligence}}
Artificial Intelligence and, particularly, of neural networks became relevant in this period as well. A lot of the important work on neural networks occurred in the 1980’s and in the 1990’s, however at that time computers did not have enough computational power. Data-sets were not big enough to be useful in practical applications. As a result, neural networks practically disappeared in the first decade of the 21st century. However, starting from 2009, neural networks gained popularity and started delivering good results in the fields of computer vision (2012) \cite{krizhevsky2012imagenet} or machine translation (2014) \cite{sutskever2014sequence}. During the last years we have seen how these techniques have been translated to robotics for tasks such as robotic grasping (2016) \cite{DBLP:journals/corr/LevinePKQ16}. In the coming years it is expected to see more innovations and these AI techniques will have high impact in robotics.

    \subsubsection{\textbf{Real-time communication solutions}}
\begin{figure}[h!]
    \centering
    \includegraphics[width=0.5\textwidth]{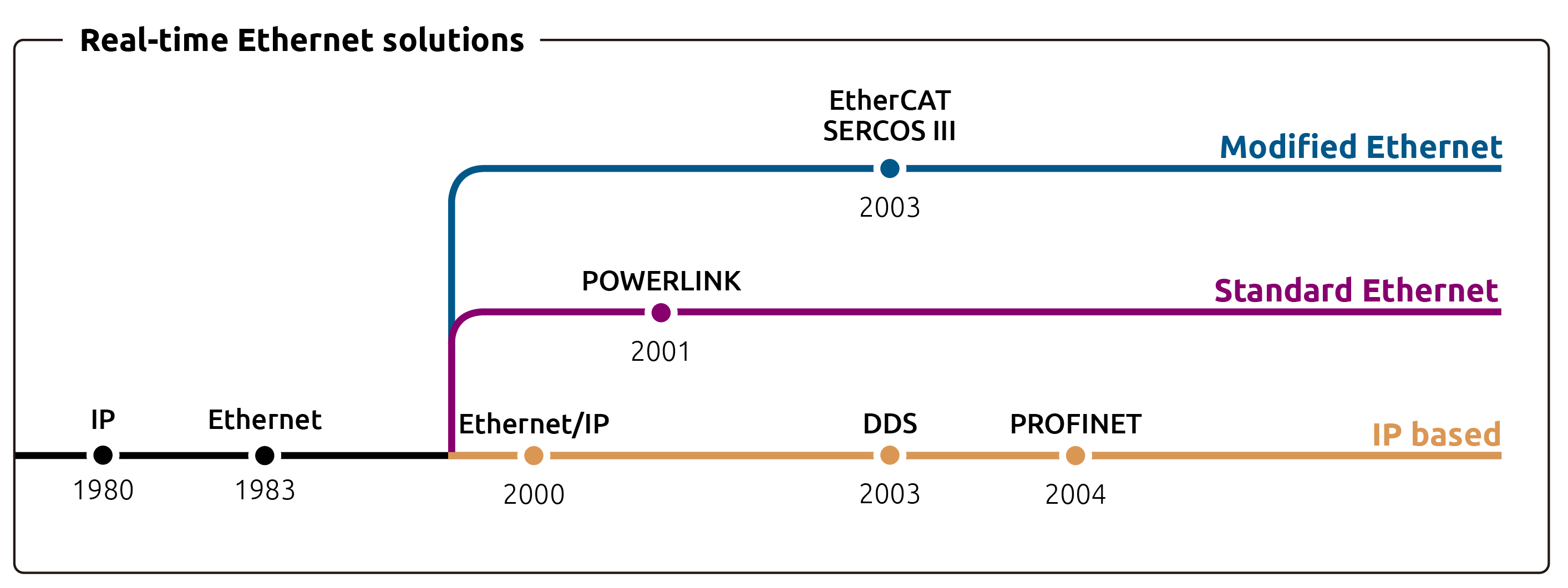}
    \caption{ Comparison between different real-time, Ethernet-based fieldbuses. As shown in this statistics EtherCAT is the most common automation communication standard. \emph{Original source} \cite{5-rt-ethernet-solutions-compared}}
    \label{Real-time}
    \end{figure}

    According to \cite{5-rt-ethernet-solutions-compared}, since 2001 reputable industrial titans introduced different Industrial real-time Ethernet standards (EtherCAT, SERCOS, PROFINET, Ethernet/IP and Ethernet PowerLink) represented in Figure \ref{Real-time}. These solutions have been widely used to criticize the deterministic and real-time capabilities of traditional Ethernet. However they do not serve to standardize the communications because the Industrial robots need to adapt and speak the factory language that have been chosen based on one of these technologies. This leads to incompatibility between different robotic or automation systems, therefore making integration task cumbersome.\\

\subsubsection{\textbf{A common infrastructure for robotics}} 
An interesting approach would be to have manufacturers agree on a common infrastructure. Such an infrastructure could define a set of electrical and logical interfaces (leaving the mechanical ones aside due to the variability of robots) that would allow industrial robot companies to produce robots and components that could interoperate, be exchanged and eventually enter into new markets. This would also lead to a competing environment where manufacturers will need to demonstrate features rather than the typical obscured environment where only some are allowed to participate.\\

\begin{tcolorbox}
Integration effort was identified as one of the main issues within robotics and particularly related to robots operating in industry. A common infrastructure typically reduces the integration effort by facilitating an environment where components can simply be connected and interoperate. Each of the infrastructure-supported components are optimized for such integration at their conception and the infrastructure handles the integration effort. At that point, components could come from different manufacturers, yet when supported by a common infrastructure, they will interoperate:

\begin{center}
\textbf{A hardware/software standardization is needed}
\end{center}

\begin{center}
\emph{"Currently, most of the time is spent dealing with the hardware/software interfaces and much less is put into behaviour development or real-world scenarios"}\\
\end{center}

\end{tcolorbox}

\noindent For robots to enter new and different fields, it seems reasonable to accept that these robots will need to adapt to the environment itself. For industrial robotics, robots have to be fluent with factory languages (EtherCAT, SERCOS, PROFINET,  Ethernet/IP and Ethernet PowerLink). One could argue that the principle is valid for service robots (e.g. households robots that will need to adapt to dish washers, washing machines, or media servers), medical robots and many other areas of robotics. Such reasoning led to the creation of the Hardware Robot Operating System (H-ROS), a vendor-agnostic hardware and software infrastructure for the creation of robot components that interoperate and can be exchanged between robots.

\begin{figure}[!h]
\centering
\includegraphics[width=0.30\textwidth]{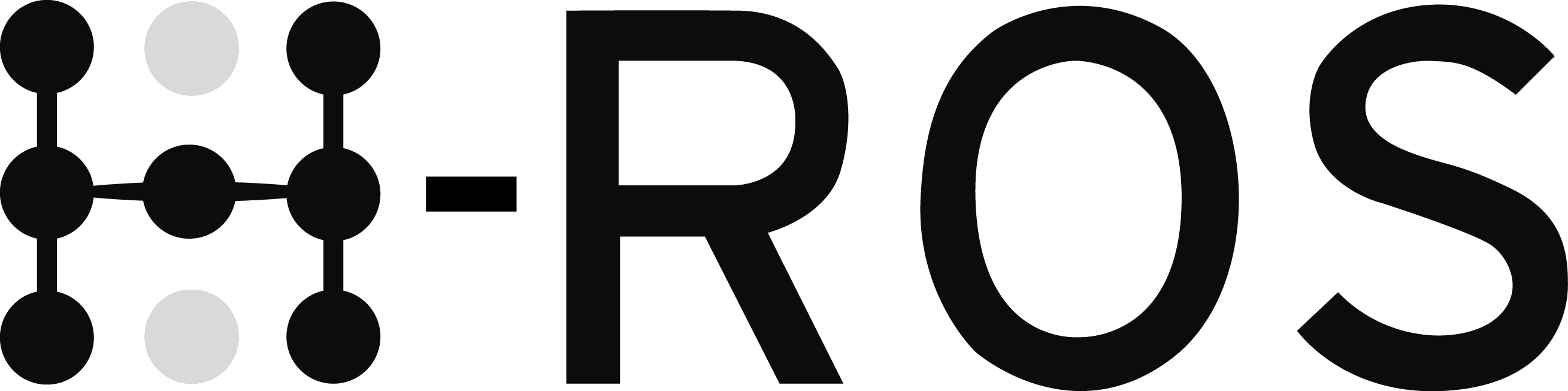}
\caption{Official logo of the Hardware Robot Operating System (H-ROS). The image was inspired by the initial logo of the Robot Operating System (ROS) with the consent of the ROS-community steered by the Open Source Robotics Foundation (OSRF).}
\label{hros_logo}
\end{figure}


\noindent H-ROS is built on top of ROS, which is the \emph{de-facto} standard for robot software development \cite{quigley2009ros}. In more detail, H-ROS utilizes the functionality of ROS 2; a redesigned version of the robot middleware that targets use cases not included in the initial design of ROS.\\

\noindent The philosophy behind H-ROS is that creating robots should be about placing together components that are compliant with the standardized H-ROS interfaces, regardless of the manufacturer. H-ROS aims to facilitate a fast way of building robots choosing the best component for each use-case from a common robot marketplace. It complies with different environment requirements (industrial, professional, personal, and others) where variables such as time constraints are critical. Building or extending robots is simplified to the point of placing H-ROS compliant components together. The user simply needs to program the cognition part (in other words, the \emph{brain}) of the robot and develop their own use-cases without facing the complexity of integrating different technologies and hardware interfaces.\\

\section{Envisioning the future of robotics}
\label{h_ros}

In order to predict the future of robotics we have analyzed the historical growth of robotics, divided into the following markets: \textbf{industrial robots}, \textbf{professional robots} and \textbf{consumer robots}.

\begin{figure}[h!]
\centering
\includegraphics[width=0.5\textwidth]{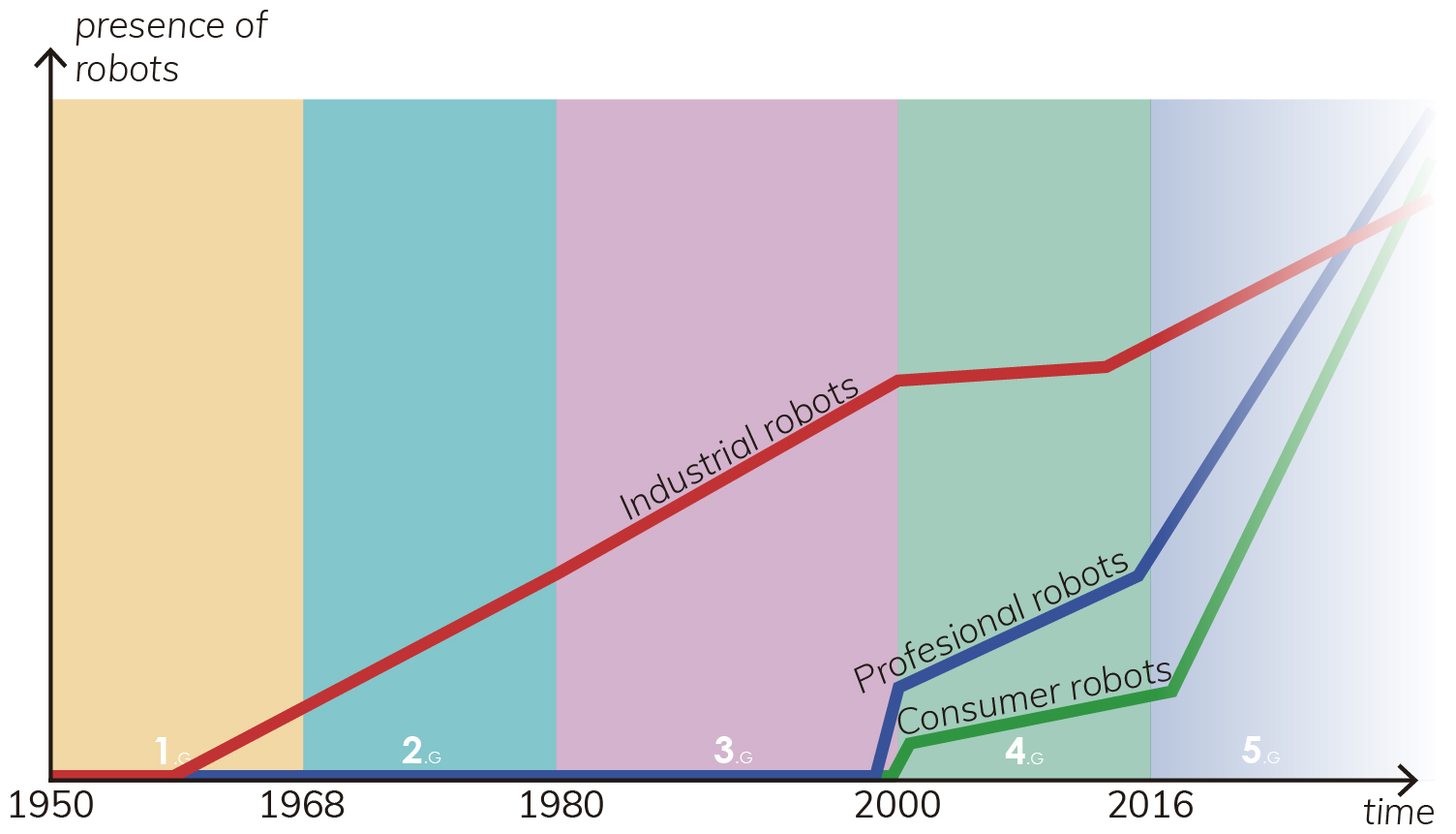}
\caption{Summary and predictability of the distribution of robots in the world, divided into three main markets: industrial robots, professional robots and consumer or personal robots.}
\label{robotpresence}
\end{figure}

\noindent The presence of industrial robots led the growth of robotics since its beginning, as shown in Figure \ref{robotpresence}, however, since 2000, the use and aim of many  robot companies and initiatives changed. A relevant amount of resources were invested into getting robots outside of industrial environments. We can distinguish two periods:

\begin{itemize}
    \item \textbf{1960-2000:} boom of the automotive industry and increased interest in industrial robots. Many started including robots in their factories to increase productivity.
    \item \textbf{Since 2000:} robot development and innovations pivot towards the consumer and professional markets which opens the door for faster innovation cycles.
\end{itemize}

\noindent Figure \ref{robotinterest} illustrates the interest in robotics obtained from a joint review of publications, conferences and events, solutions and corporations:

\begin{figure}[t]
\centering
\includegraphics[width=0.5\textwidth]{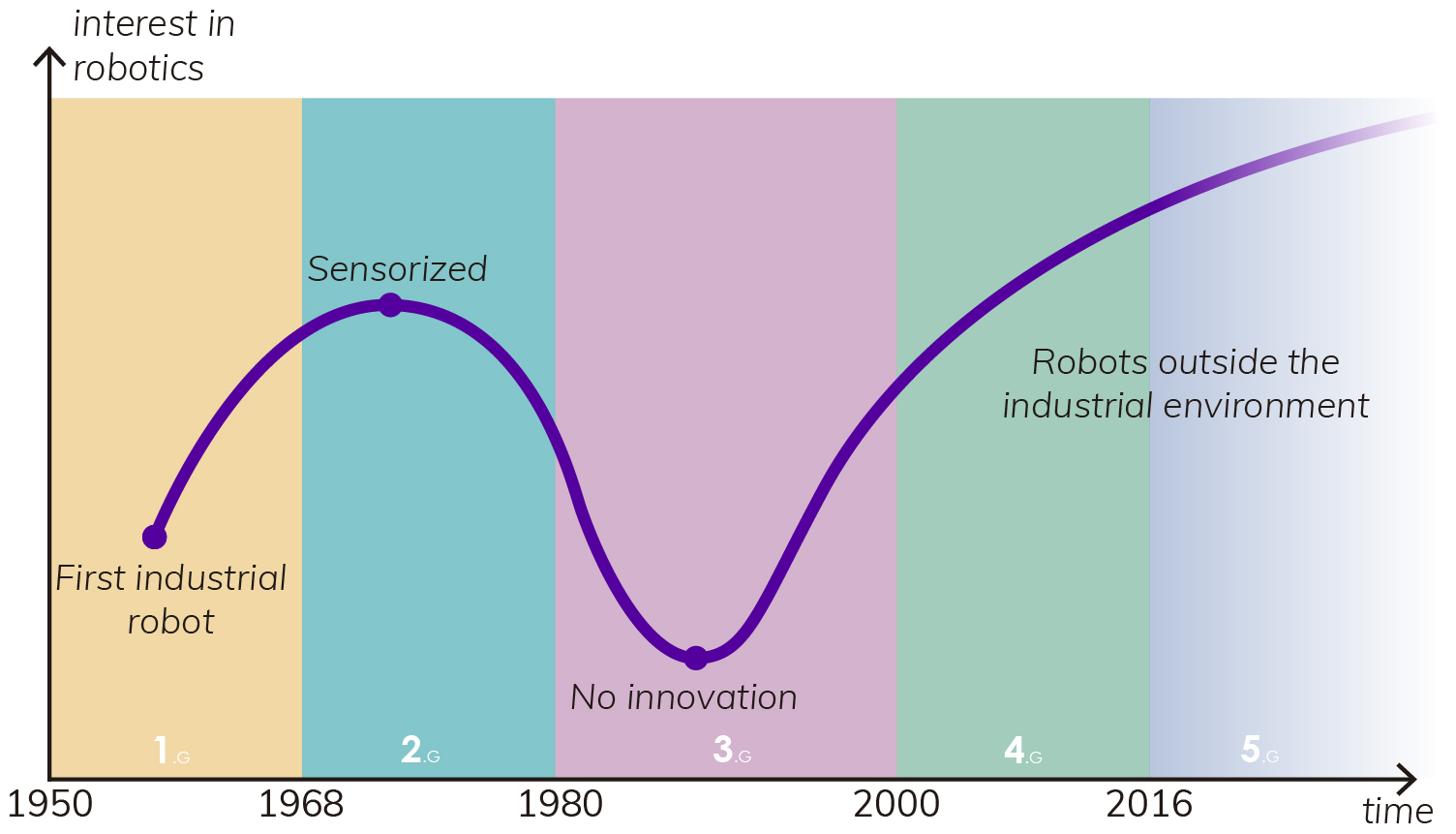}
\caption{The general interest in robotics since its inception obtained from a joint review of publications, conferences and events, solutions and corporations.}
\label{robotinterest}
\end{figure}

\subsection*{\textbf{Generation 5: Collaborative and personal robots}}

\begin{tcolorbox}[colback=gray!10]
\textbf{Characteristics:}\\
\begin{itemize}
    \item Robots and humans share same environment and collaborate.
    \item Reconfigurable robots.
    \item Robots help humans enhance every-day activities.
    \item Modular robots and components.\\    
\end{itemize}
\end{tcolorbox}

With latest developments in AI (e.g. AlphaGo beats the world-class player in Go \cite{7515285}, an achievement that was not expected for many years) being translated to robotics and the recent investments in the field, there is a high expectation for the future development of robotics. Our team believes that the next generation robots will reflect all the technological advances that developers and researchers have made in recent years.\\
\newline
According to the characteristics of the 5$^{th}$ generation, robots will be able to coexist with the humans, enhance human capabilities, simplify and improve life. We foresee a boom in collaborative and personal robots:

\begin{figure}[H]
\centering
\includegraphics[width=0.4\textwidth]{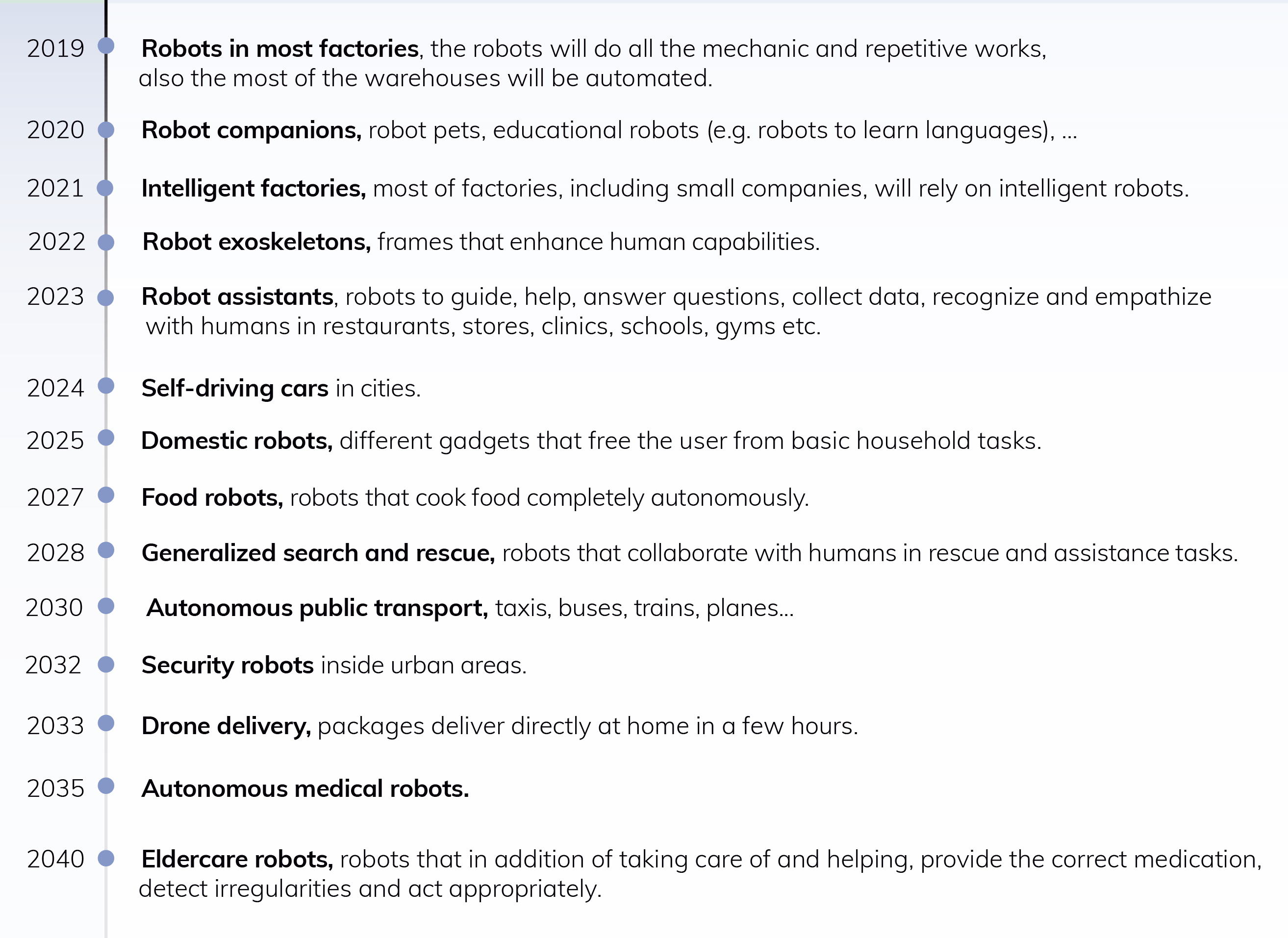}
\caption{Envisioning the future robots.}
\label{future}
\end{figure}

\noindent The list presented in the Figure \ref{future} presents our perception regarding the near future of robotics by taking into account its technical feasibility.


\section{Discussion and conclusions}
\label{discussion}
Robotics did not grow as much as expected. Visionaries predicted that by 2020 robots would be in our every day lives helping with daily tasks. To the best of our knowledge, this fact is highly unlikely, however the results obtained through this research unveiled a promising future for the field. The importance of robotics and its potential is being intensely highlighted and a number of standardization bodies are taking steps towards creating a common set of rules to govern the interaction with these machines. \\
The following subsections reflect some of the most relevant conclusions taken from our review:


\subsection {Lack of compatible systems}

\begin{itemize}

\item \textbf{Hardware/software incompatibility:} Building a new robot is a bumpy road. The incompatibility between components/elements/languages, take significant time and effort leading to the fact that during development of a new robotic solution most of the time  is dedicated in the integration and not to the behavior or developing innovative solutions.

\item \textbf{Different programming languages and environments:} In an attempt to dominate the market the main industrial robots manufacturers had created their own programming languages to keep their technology as closed as possible. As consequence there are many incompatible components/robots and integration between components from different manufactures is very cumbersome, time consuming process which sometimes is not even feasible. In order to "simplify" integration effort many users decide to bound to solutions provided from a single manufacturer which increases the overall costs.

\end{itemize}
 
\begin{tcolorbox}

\begin{center}
\textbf{A common infrastructure for robot components is needed, H-ROS to lead the change.}\\
\end{center}

\noindent Making robotics hardware more affordable, versatile, and “standardized” is hugely important for the field, as Aaron Dollar, Francesco Mondada, Alberto Rodriguez, and Giorgio Metta, who guest edited the special issue \cite{dollar2017open}:

\begin{center}
\emph{"In the field of robotics, there has existed a relatively large void in terms of the availability of adequate hardware, particularly for research applications. The few systems that have been appropriate for advanced applications have been extremely costly and not very durable. For those and other reasons, innovation in commercially available hardware is extremely slow, with a historically small market and expensive and slow development cycles. Effective open source hardware that can be easily and inexpensively fabricated would not only substantially lower costs and increase accessibility to these systems, but would drastically improve innovation and customization of available hardware."}
\end{center}
\end{tcolorbox}
 
\begin{itemize} 

\item \textbf{The Industrial robot industry —--will it remain only a supplier industry?---} For  some,  the  industrial  robot  industry  is  a  supplier industry. It supplies components and systems to larger  industries,  mainly,  the  manufacturing  industry. These  groups argue that the manufacturing industry is dominated by the PLC, motion control and communication suppliers which together with the big customers are setting the standards to capitalize on the cost savings from Ethernet by extending their own standard to include Ethernet. In doing so, their customers receive some of the benefits from Ethernet but are still locked into the proprietary networks for the long term. Frequently forgotten in these discussions is the fact that it is not just technical properties such as performance and transfer rates that count, it is also the soft facts like ease of implementation, openness, vendor independence, risk avoidance, conformity, interoperability, long-term availability, and overall distribution that makes a standard gain acceptance and even thrive. The Industrial robots need to adapt and speak factory language (such as, PROFINET, ETHERCAT, Modbus TCP, Ethernet/IP, CANOPEN, DEVICENET) which for each factory, might be different. As a result, most robotic peripheral manufacturers suffer from supporting many different protocols which requires a lot of development time that does not add functionality to the product.

\item \textbf{Competing by obscuring is slowing industry}\\ The close attitude of most industrial robot companies is typically justified by the existing competition in this environment. Such attitude leads to a lack of understanding between different manufacturers and solutions but in exchange, some believe that it secures clients and favours competition. Our results indicate that this behavior is slowing progress, innovation and new solutions in the field of industrial robots.

\end{itemize}

    

\subsection{The hype cycle of robotics}

\begin{figure}[h!]
\centering
 \includegraphics[width=0.5\textwidth]{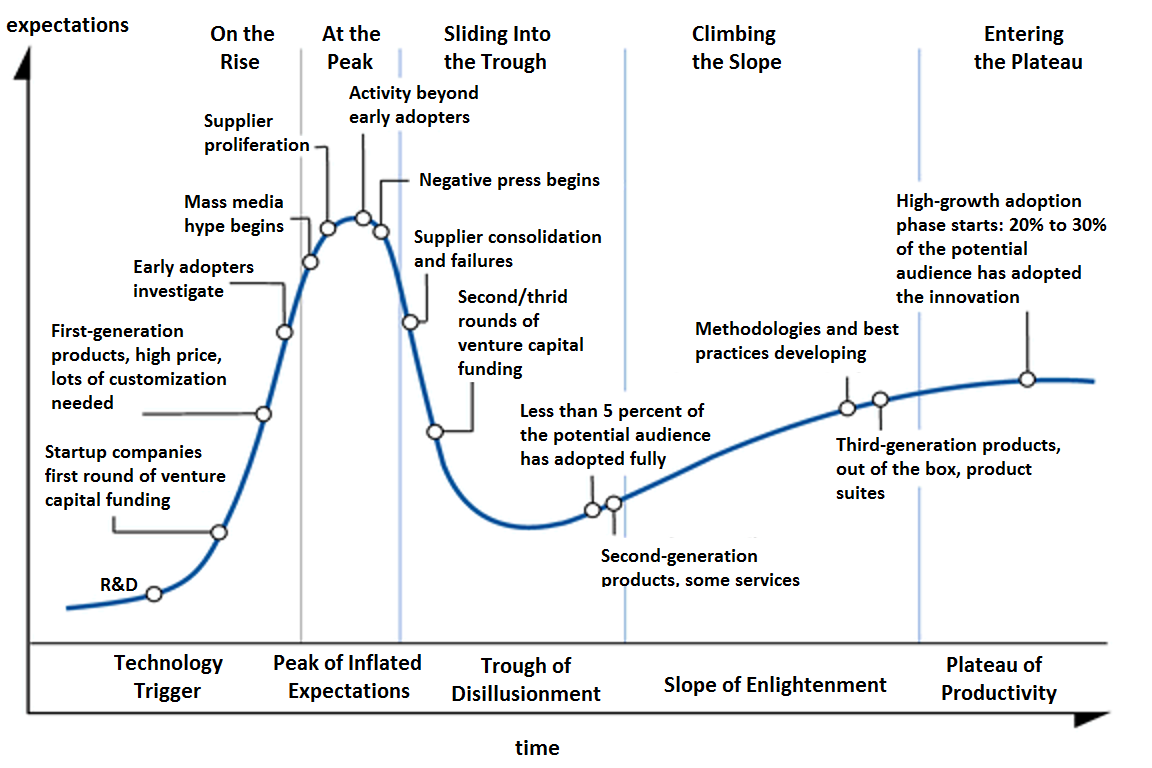}
\caption{\emph{Gartner's Hype Cycles} offer an overview of relative maturity of technologies in a certain domain. They provide not only a scorecard to separate hype from reality, but also models that help enterprises decide when they should adopt a new technology. \cite{linden2003understanding}. }
\label{hype}
\end{figure}

Robotics, like many other technologies, suffered from an inflated set of expectations which resulted in a decrease of the developments and results during the 1990s. Figure \ref{robotinterest} pictured the evolution of the general interest in robotics. Such graph displays a well known trajectory typically known as the \emph{hype cycle}\cite{linden2003understanding}. Figure \ref{hype} pictures the hype cycle as defined by Linden and Fenn. The graph illustrates the five different states that a technology goes through before entering mass-adoption: a) technology trigger, b) peak of inflated expectations, c) trough of disillusionment, d) slope of enlightenment and e) plateau of productivity.

Comparing Figures \ref{robotinterest} and \ref{hype} we  conclude that the state of robotics is past the "second-generation products, some services" and somewhere within the \emph{slope of enlightenment}. 



\begin{tcolorbox}

\begin{center}
\textbf{Industrial robots on the rise}\\
\end{center}

\noindent Initiatives such as the U.S. Advanced Robotics for Manufacturing (ARM) Institute are pushing the boundaries of industrial robots once again \cite{armInstitute}. Collaborative robots are increasingly improving the performance of the manufacturing processes while improving the interaction with humans.\\ 

\end{tcolorbox}



\begin{tcolorbox}
\begin{center}
\textbf{Robots everywhere}\\
\end{center}

\noindent Robots and automation are actively being introduced in many disciplines. With the growing popularity of such systems, we observe a transition that goes from \textbf{mass manufacturing to a mass customization}, particularly for industrial robots. Increasingly, personal (e.g. cleaning robots) and professional robots (e.g. service robots) are also demanding such customizations. We foresee that an infrastructure such as H-ROS will favor the creation of modular robots whose components could be easily exchanged or replaced meeting the growing needs for customization. \\


\end{tcolorbox}

\subsection{Artificial Intelligence (AI) taking over:} 

Current robot systems are designed by teams with multi-disciplinary skills. The design of the control mechanisms is one of the critical tasks. Typically, the traditional approach to design such systems require going from observations to final control commands through a) state estimation, b) modeling and prediction, c) planning and d) low level control (for example, inverse kinematics). This whole process requires fine tuning every step of the funnel incurring into a relevant complexity where optimization at every step have a direct impact in the final result.\\
\newline
The work of Levine et al. \cite{DBLP:journals/corr/LevinePKQ16} shows a promising path towards simplifying the construction of robot behaviors through the use of deep neural networks as a replacement of the whole funnel described above\footnote{The use of neural networks could also replace individual tasks within the funnel.}. Our team envisions that the use of deep neural networks will become of great relevance in the field of robotics in the coming years. These abstractions will empower roboticists to train robot models in specific applications (end-to-end) empowering engineers and robots to tackle more complex problems.




\hfill

\noindent Robots are going to change the society, in the same way computers changed our life. The introduction of robotics in our society will be disruptive. Work, communication, transportation and even social life will be affected by robotics. This will lead to change from \textbf{mass customization to mass integration}, so that robots and humans coexist helping each other.\\

\section*{Acknowledgment}

This review was funded and supported by Acutronic Robotics\footnote{www.acutronicrobotics.com}, a firm focused on the development of next-generation robot solutions for a range of clients.\\

\ifCLASSOPTIONcaptionsoff
  \newpage
\fi

\bibliographystyle{IEEEtran}
\bibliography{references}

\begin{thebibliography}{10}
\providecommand{\url}[1]{#1}
\csname url@samestyle\endcsname
\providecommand{\newblock}{\relax}
\providecommand{\bibinfo}[2]{#2}
\providecommand{\BIBentrySTDinterwordspacing}{\spaceskip=0pt\relax}
\providecommand{\BIBentryALTinterwordstretchfactor}{4}
\providecommand{\BIBentryALTinterwordspacing}{\spaceskip=\fontdimen2\font plus
\BIBentryALTinterwordstretchfactor\fontdimen3\font minus
  \fontdimen4\font\relax}
\providecommand{\BIBforeignlanguage}[2]{{%
\expandafter\ifx\csname l@#1\endcsname\relax
\typeout{** WARNING: IEEEtran.bst: No hyphenation pattern has been}%
\typeout{** loaded for the language `#1'. Using the pattern for}%
\typeout{** the default language instead.}%
\else
\language=\csname l@#1\endcsname
\fi
#2}}
\providecommand{\BIBdecl}{\relax}
\BIBdecl

\bibitem{low2007industrial}
\BIBentryALTinterwordspacing
K.~Low, \emph{Industrial Robotics: Programming, Simulation and Applications},
  ser. ARS, Advanced robotic systems international.\hskip 1em plus 0.5em minus
  0.4em\relax Pro-Literatur-Verlag, 2007. [Online]. Available:
  \url{https://books.google.es/books?id=xj37AX\_VqHYC}
\BIBentrySTDinterwordspacing

\bibitem{erleEnvisioning}
\BIBentryALTinterwordspacing
E.~Robotics, ``Envisioning robotics,'' \emph{Official Webpage}, April 2017.
  [Online]. Available: \url{http://erlerobotics.com/envisioning-robotics/}
\BIBentrySTDinterwordspacing

\bibitem{transistor}
P.~K. Bondyopadhyay, ``In the beginning [junction transistor],''
  \emph{Proceedings of the IEEE}, vol.~86, no.~1, pp. 63--77, Jan 1998.

\bibitem{506395}
A.~E. Bryson, ``Optimal control-1950 to 1985,'' \emph{IEEE Control Systems},
  vol.~16, no.~3, pp. 26--33, Jun 1996.

\bibitem{middleditch1973survey}
A.~E. Middleditch, \emph{Survey of numerical controller technology}.\hskip 1em
  plus 0.5em minus 0.4em\relax Production Automation Project, University of
  Rochester, 1973.

\bibitem{saxena2009invention}
\BIBentryALTinterwordspacing
A.~Saxena, \emph{Invention of Integrated Circuits: Untold Important Facts},
  ser. International series on advances in solid state electronics and
  technology.\hskip 1em plus 0.5em minus 0.4em\relax World Scientific, 2009.
  [Online]. Available: \url{https://books.google.es/books?id=-3lpDQAAQBAJ}
\BIBentrySTDinterwordspacing

\bibitem{firsthumanoid}
B.~Gates, ``A robot in every home,'' \emph{Scientific American}, vol. 296, pp.
  58--65, 2007.

\bibitem{punchcard}
B.~Trikha, ``A journey from floppy disk to cloud storage,'' \emph{International
  Journal on Computer Science and Engineering}, vol.~2, no.~4, pp. 1449--1452,
  2010.

\bibitem{colossus}
B.~Copeland, ``Colossus: its origins and originators,'' \emph{IEEE Annals of
  the History of Computing}, vol.~26, no.~04, pp. 38--45, 2004.

\bibitem{unimate}
\BIBentryALTinterwordspacing
P.~Mickle, ``A peep into the automated future,'' \emph{The Trentonian}, 1961.
  [Online]. Available: \url{http://www.capitalcentury.com/1961.html}
\BIBentrySTDinterwordspacing

\bibitem{shakey}
M.~V. Georges~Giralt, Raja~Chatila, ``An integrated navigation and motion
  control system for autonomous multisensory mobile robots,'' in
  \emph{Autonomous Robot Vehicles}, vol.~7.\hskip 1em plus 0.5em minus
  0.4em\relax Springer New York, 1990, pp. 420--443.

\bibitem{bryan1988programmable}
L.~Bryan and E.~Bryan, ``Programmable controllers,'' 1988.

\bibitem{kuka}
S.~Shepherd and A.~Buchstab, ``Kuka robots on-site,'' in \emph{Robotic
  Fabrication in Architecture, Art and Design 2014}.\hskip 1em plus 0.5em minus
  0.4em\relax Springer, 2014, pp. 373--380.

\bibitem{cutkosky1982position}
M.~R. Cutkosky and P.~K. Wright, ``Position sensing wrists for industrial
  manipulators.'' DTIC Document, Tech. Rep., 1982.

\bibitem{roboticsera}
J.~Maeda, ``Current research and development and approach to future automated
  construction in japan,'' \emph{Construction Research Congress 2005:
  Broadening Perspectives}, pp. 1--11, 2005.

\bibitem{internet}
M.~Castells, ``Lessons from the history of the intenet,'' in \emph{The Internet
  galaxy: Reflections on the Internet, business, and society}.\hskip 1em plus
  0.5em minus 0.4em\relax Oxford University Press, 2002, pp. 20--33.

\bibitem{ethernet}
S.~Yu, ``Ieee 802.3™ 'standard for ethernet' marks 30 years of innovation and
  global market growth,'' \emph{IEEE Standards Association}, 2013.

\bibitem{linux}
M.~C. Daniel P.~Bovet, ``Understanding the linux kernel: from i/o ports to
  process management,'' vol.~3.\hskip 1em plus 0.5em minus 0.4em\relax
  O’Reilly Media, 2005.

\bibitem{linuxreal}
M.~Barabanov, ``A linux--based real--time operating system,'' New Mexico
  Institute of Mining and Technology, Tech. Rep., 1997.

\bibitem{Yodaiken_cheapoperating}
V.~Yodaiken, ``Cheap operating systems research and teaching with linux,''
  Tech. Rep., 1996.

\bibitem{val}
\BIBentryALTinterwordspacing
K.~Srihari and M.~P. Deisenroth, \emph{Robot Programming Languages---A State of
  the Art Survey}.\hskip 1em plus 0.5em minus 0.4em\relax Berlin, Heidelberg:
  Springer Berlin Heidelberg, 1988, pp. 625--635. [Online]. Available:
  \url{http://dx.doi.org/10.1007/978-3-642-73890-6_76}
\BIBentrySTDinterwordspacing

\bibitem{val1}
W.~A. Gruver, B.~I. Soroka, J.~J. Craig, and T.~L. Turner, ``Industrial robot
  programming languages: A comparative evaluation,'' \emph{IEEE Transactions on
  Systems, Man, and Cybernetics}, vol. SMC-14, no.~4, pp. 565--570, 1984.

\bibitem{fanuc}
\BIBentryALTinterwordspacing
J.~Lapham, ``Robotscript™: the introduction of a universal robot programming
  language,'' \emph{Industrial Robot: An International Journal}, vol.~26,
  no.~1, pp. 17--25, 1999. [Online]. Available:
  \url{http://dx.doi.org/10.1108/01439919910250188}
\BIBentrySTDinterwordspacing

\bibitem{abb}
\BIBentryALTinterwordspacing
J.~Hollingum, ``Abb focus on “lean robotization”,'' \emph{Industrial Robot:
  An International Journal}, vol.~21, no.~4, pp. 15--16, 1994. [Online].
  Available: \url{http://dx.doi.org/10.1108/01439919410068140}
\BIBentrySTDinterwordspacing

\bibitem{LEGO}
D.~C. Cliburn, ``Experiences with the lego mindstorms throughout the
  undergraduate computer science curriculum,'' in \emph{Proceedings. Frontiers
  in Education. 36th Annual Conference}, Oct 2006, pp. 1--6.

\bibitem{AIBO}
M.~Fujita, ``On activating human communications with pet-type robot aibo,''
  \emph{Proceedings of the IEEE}, vol.~92, no.~11, pp. 1804--1813, Nov 2004.

\bibitem{jones2005autonomous}
\BIBentryALTinterwordspacing
J.~Jones, N.~Mack, D.~Nugent, and P.~Sandin, ``Autonomous floor-cleaning
  robot,'' Apr.~26 2005, uS Patent 6,883,201. [Online]. Available:
  \url{https://www.google.com/patents/US6883201}
\BIBentrySTDinterwordspacing

\bibitem{yumi}
\BIBentryALTinterwordspacing
A.~group, ``Abb introduces yumi®, world’s first truly collaborative dual-arm
  robot,'' \emph{ABB Press release}, 2015. [Online]. Available:
  \url{http://www04.abb.com/global/seitp/seitp202.nsf/0/5869f389ad26c612c1257e26001c974c/$file/15_23+GPR+YuMi+Hannover+pr.pdf}
\BIBentrySTDinterwordspacing

\bibitem{player}
\BIBentryALTinterwordspacing
A.~H. Brian P.~Gerkey, Richard T.~Vaughan, ``The player/stage project: Tools
  for multi-robot and distributed sensor systems,'' \emph{11th International
  Conference on Advanced Robotics (ICAR 2003)}, vol.~1, pp. 317--323, 2003.
  [Online]. Available:
  \url{http://robotics.usc.edu/~gerkey/research/final_papers/icar03-player.pdf}
\BIBentrySTDinterwordspacing

\bibitem{gazebo}
N.~Koenig and A.~Howard, ``Design and use paradigms for gazebo, an open-source
  multi-robot simulator,'' in \emph{2004 IEEE/RSJ International Conference on
  Intelligent Robots and Systems (IROS) (IEEE Cat. No.04CH37566)}, vol.~3,
  2004, pp. 2149--2154.

\bibitem{ROS}
\BIBentryALTinterwordspacing
W.~Garage, ``Robot operating system,'' 2009. [Online]. Available: \url{Robot
  operating system}
\BIBentrySTDinterwordspacing

\bibitem{RPi}
\BIBentryALTinterwordspacing
M.~Richardson and S.~Wallace, \emph{Getting Started with Raspberry Pi}, ser.
  EBSCOhost ebooks online.\hskip 1em plus 0.5em minus 0.4em\relax O'Reilly
  Media, 2012. [Online]. Available:
  \url{https://books.google.es/books?id=xYhMlilTwC4C}
\BIBentrySTDinterwordspacing

\bibitem{krizhevsky2012imagenet}
A.~Krizhevsky, I.~Sutskever, and G.~E. Hinton, ``Imagenet classification with
  deep convolutional neural networks,'' in \emph{Advances in neural information
  processing systems}, 2012, pp. 1097--1105.

\bibitem{sutskever2014sequence}
I.~Sutskever, O.~Vinyals, and Q.~V. Le, ``Sequence to sequence learning with
  neural networks,'' in \emph{Advances in neural information processing
  systems}, 2014, pp. 3104--3112.

\bibitem{DBLP:journals/corr/LevinePKQ16}
\BIBentryALTinterwordspacing
S.~Levine, P.~Pastor, A.~Krizhevsky, and D.~Quillen, ``Learning hand-eye
  coordination for robotic grasping with deep learning and large-scale data
  collection,'' \emph{CoRR}, vol. abs/1603.02199, 2016. [Online]. Available:
  \url{http://arxiv.org/abs/1603.02199}
\BIBentrySTDinterwordspacing

\bibitem{5-rt-ethernet-solutions-compared}
Kingstar, ``White paper: 5 real-time, ethernet-based fieldbuses compared,''
  Kingstar, Tech. Rep., January 2016.

\bibitem{quigley2009ros}
M.~Quigley, K.~Conley, B.~Gerkey, J.~Faust, T.~Foote, J.~Leibs, R.~Wheeler, and
  A.~Y. Ng, ``Ros: an open-source robot operating system,'' in \emph{ICRA
  workshop on open source software}, vol.~3, no. 3.2.\hskip 1em plus 0.5em
  minus 0.4em\relax Kobe, 2009, p.~5.

\bibitem{7515285}
C.~S. Lee, M.~H. Wang, S.~J. Yen, T.~H. Wei, I.~C. Wu, P.~C. Chou, C.~H. Chou,
  M.~W. Wang, and T.~H. Yan, ``Human vs. computer go: Review and prospect
  [discussion forum],'' \emph{IEEE Computational Intelligence Magazine},
  vol.~11, no.~3, pp. 67--72, Aug 2016.

\bibitem{dollar2017open}
A.~Dollar, F.~Mondada, A.~Rodriguez, and G.~Metta, ``Open-source and widely
  disseminated robot hardware [from the guest editors],'' \emph{IEEE Robotics
  \& Automation Magazine}, vol.~24, no.~1, pp. 30--31, 2017.

\bibitem{linden2003understanding}
A.~Linden and J.~Fenn, ``Understanding gartner’s hype cycles,''
  \emph{Strategic Analysis Report N{\textordmasculine} R-20-1971. Gartner,
  Inc}, 2003.

\bibitem{armInstitute}
\BIBentryALTinterwordspacing
A.~institute, ``Advanced robotics for manufacturing institute,'' \emph{Official
  Webpage}, April 2017. [Online]. Available: \url{http://www.arminstitute.org/}
\BIBentrySTDinterwordspacing

\end{thebibliography}

\end{document}